\documentclass[journal]{IEEEtran}
\def\BibTeX{{\rm B\kern-.05em{\sc i\kern-.025em b}\kern-.08em
T\kern-.1667em\lower.7ex\hbox{E}\kern-.125emX}}
\usepackage[english]{babel}
\usepackage[noadjust]{cite}
\usepackage{algorithm}
\usepackage{algpseudocode}
\usepackage{booktabs}
\usepackage[table,x11names]{xcolor}
\usepackage{siunitx}
\DeclareSIUnit{\dBm}{\deci\belmilliwatt}
\DeclareSIUnit{\dBi}{\deci\beli}
\usepackage[table]{xcolor} 
\definecolor{lightergray}{rgb}{0.9, 0.9, 0.9}
\usepackage{titlesec}
\titlespacing*{\section}{0pt}{0.3\baselineskip}{0.2\baselineskip}
\titlespacing*{\subsection}{0pt}{0.3\baselineskip}{0.2\baselineskip}
\ifCLASSINFOpdf
\else
   \usepackage[dvips]{graphicx}
\fi
\usepackage{url}
\usepackage[scaled]{helvet}

\usepackage{multirow}
\usepackage{tabularx}
\usepackage{graphicx}
\usepackage{optidef}
\usepackage{amsmath}
\usepackage{bm}
\usepackage{amssymb}
\usepackage{xcolor}
\usepackage{caption}
\usepackage{subcaption}
\usepackage{booktabs}
\usepackage{lipsum}%
\usepackage{siunitx}    
\usepackage[acronym]{glossaries}
\usepackage{cleveref}

\usepackage[binary-units]{siunitx}
\DeclareSIUnit{\belmilliwatt}{Bm}
\DeclareSIUnit{\dBm}{\deci\belmilliwatt}
\DeclareSIUnit{\belisotropic}{Bi}
\DeclareSIUnit{\dBi}{\deci\belisotropic}
\DeclareSIUnit{\foot}{ft}

\makeglossaries

\newacronym{itm}{ITM}{irregular terrain model}

\newacronym{ehata}{eHata}{extended Hata}

\newacronym{us}{US}{United States}

\newacronym{id}{ID}{identifier}

\newacronym{rsrp}{RSRP}{Reference Signal Received Power}

\newacronym{gps}{GPS}{Global Positioning System}
\newacronym{earfcn}{EARFCN}{E-UTRA Absolute Radio Frequency Channel Number}

\newacronym{lsf}{LSF}{large-scale fading}

\newacronym{aoa}{AoA}{angle of arrival}

\newacronym{aod}{AoD}{angle of departure}
\newacronym{bs}{BS}{base stations}

\newacronym{ml}{ML}{machine learning}
\newacronym{mae}{MAE}{mean absolute error}

\newacronym{los}{LoS}{line-of-sight}
\newacronym{nlos}{nLoS}{non-LoS}
\newacronym{dsm}{DSM}{Digital Surface Model}
\newacronym{dhm}{NDHM}{Normalized Digital Height Model}

\newacronym{qos}{QoS}{quality of service}

\usepackage{atbegshi,picture}

\AtBeginShipout{\AtBeginShipoutUpperLeft{%
  \put(\dimexpr\paperwidth-1cm\relax,-1.5cm){\makebox[0pt][r]{\framebox{This manuscript has been accepted for publication in ICC 2024. The final content may differ}}}%
}}

\begin{document}
\bstctlcite{BSTcontrol}
 \title{Simulation-Enhanced Data Augmentation for Machine Learning Pathloss Prediction}


\author{
\IEEEauthorblockN{Ahmed P. Mohamed\IEEEauthorrefmark{1}, Byunghyun Lee\IEEEauthorrefmark{1}, Yaguang Zhang\IEEEauthorrefmark{1}, Max Hollingsworth\IEEEauthorrefmark{2}, C. Robert Anderson\IEEEauthorrefmark{3},
\\James V. Krogmeier\IEEEauthorrefmark{1}, David J. Love\IEEEauthorrefmark{1}} 

\IEEEauthorblockA{\IEEEauthorrefmark{1}Elmore Family School of Electrical and Computer Engineering, Purdue University, West Lafayette, IN 47907\\
\IEEEauthorrefmark{2}University of Colorado, Boulder, CO 80309\\
\IEEEauthorrefmark{3}Bradley Department of Electrical and Computer Engineering, Virginia Tech, Blacksburg, VA 24061\\
\IEEEauthorblockA{Email: {\{mohame23, lee4093, ygzhang, jvk, djlove\}@purdue.edu, maho5047@colorado.edu, chanders@vt.edu}}
\vspace{-2em}}

\thanks{
This work is supported by the National Science Foundation under grants EEC-1941529, CNS-2212565, and CNS-2225578. (Ahmed P. Mohamed and Byunghyun Lee contributed equally to this work.)}
}

\maketitle

\begin{abstract}

Machine learning (ML) offers a promising solution to pathloss prediction.
However, its effectiveness can be degraded by the limited availability of data.
To alleviate these challenges, this paper introduces a novel simulation-enhanced data augmentation method for \gls{ml} pathloss prediction.
Our method integrates synthetic data generated from a cellular coverage simulator and independently collected real-world datasets. These datasets were collected through an extensive measurement campaign in different environments, including farms, hilly terrains, and residential areas. This comprehensive data collection provides vital ground truth for model training. 
A set of channel features was engineered, including geographical attributes derived from LiDAR datasets. 
These features were then used to train our prediction model, incorporating the highly efficient and robust gradient boosting \gls{ml} algorithm, CatBoost. 
The integration of synthetic data, as demonstrated in our study, significantly improves the generalizability of the model in different environments, achieving a remarkable improvement of approximately \text{\bfseries\SI[detect-weight]{12}{\deci\bel}} in terms of \acrlong{mae} for the best-case scenario.
Moreover, our analysis reveals that even a small fraction of measurements added to the simulation training set, with proper data balance, can significantly enhance the model's performance.

\end{abstract}


\IEEEpeerreviewmaketitle

\section{Introduction}

\glsresetall


    Radio signals experience pathloss as they propagate to a receiver.
Pathloss refers to the attenuation of the signal of a communication link between the transmitter and the receiver.
Accurately estimating pathloss is fundamental for coverage estimation and interference analysis, which are key to effective network planning.
Additionally, accurate pathloss prediction enables improved mobility management, such as handover and quality-of-service prediction.
However, predicting pathloss is challenging because it depends on various factors, including propagation distance, geometry (e.g., buildings and trees), antenna pattern and carrier frequency \cite{phillipsSurveyWirelessPath2013}.

The fundamental model for the prediction of pathloss is the Friis equation, which calculates the loss of transmission in the free space based on the distance and carrier frequency \cite{friisNoteSimpleTransmission1946}. 
Terrain-based models improve accuracy by incorporating topological features. 
The \gls{itm}, often referred to as the Longley-Rice model, aims to predict pathloss considering terrain profiles \cite{huffordDEPARTMENTCOMMERCE}.
The \gls{ehata} model adds environment categories, such as urban, suburban or rural settings, to account for different endpoint propagation environments \cite{drocellaGHzExclusionZone}.
However, in many practical scenarios where multiple environments are combined, these models may fail to provide an accurate pathloss prediction.

Stochastic models add a random variable to a deterministic pathloss model to describe the randomness (e.g., scattering and multipath effect) in a wireless link.
The most widely used model is the log-normal shadowing model, which accounts for shadowing with a Gaussian random variable.
Furthermore, the COST 231 model \cite{COSTAction231}, 3GPP spatial channel model \cite{3gpp.38.901}, and QuadRiGa \cite{jaeckelQuaDRiGa3DMultiCell2014} empirically model the shadowing distribution.
These models are useful in that they can be easily employed without high complexity and provide a general understanding of signal propagation characteristics in various environments.
However, since these models do not consider the exact surroundings and environments of the transmitter and receiver, they may not capture site-specific features.

Ray-tracing models simulate the propagation of electromagnetic waves using deterministic modeling of reflection, diffraction, and scattering \cite{lawtonApplicationDeterministicRay1994,heDesignApplicationsHighPerformance2019}.
Ray-tracing has been used extensively for research purposes as it offers detailed information about the structure of a wireless channel, including angle of
arrival, angle of departure, and path delays.
Despite this, the computational cost of ray-tracing becomes unaffordable as the simulation scale increases, making it less suitable for large-scale implementations.

To overcome these issues, learning-based models have attracted significant research interest \cite{ratnamFadeNetDeepLearningBased2021a,vanleerImprovingPropagationModel2021,reus-munsMachineLearningbasedMmWave2022,guptaMachineLearningBasedUrban2022,zhangCellularNetworkRadio2020}.
Despite their ability to learn site-specific features, the prediction performance of learning-based models is largely based on the availability of extensive high-quality data.
Nonetheless, the collection of such datasets poses challenges as it involves considerable human labor and sophisticated measurement tools.
Although driving tests are often used to streamline this process, they are not practical in locations such as farms and school campuses, which can bias the data sets to publicly accessible roads.
Due to the lack of data, a learning-based prediction model may struggle to generalize these unfamiliar radio environments. 
This is a major issue for rural wireless communication applications, where diverse propagation environments coexist in a wide area \cite{zhangChallengesOpportunitiesFuture2021a}.

In the \gls{ml} domain, simulation-assisted data augmentation has been actively investigated due to its cost-effectiveness and convenience \cite{tang2021augmenting,mohamed2021knowledge}.
However, the use of simulation-aided data augmentation remains relatively unexplored in the domain of wireless communications, and more specifically, in pathloss prediction.
In \cite{zhangSimulationAidedMeasurementBasedChannel2020}, the authors used simulation data to cover inaccessible geographic regions for site-specific channel modeling.
However, this work focused on augmenting a partial dataset for a single geographic area rather than augmenting a dataset with different environments.



\begin{figure}
\centering
    \includegraphics[scale=0.08]
    {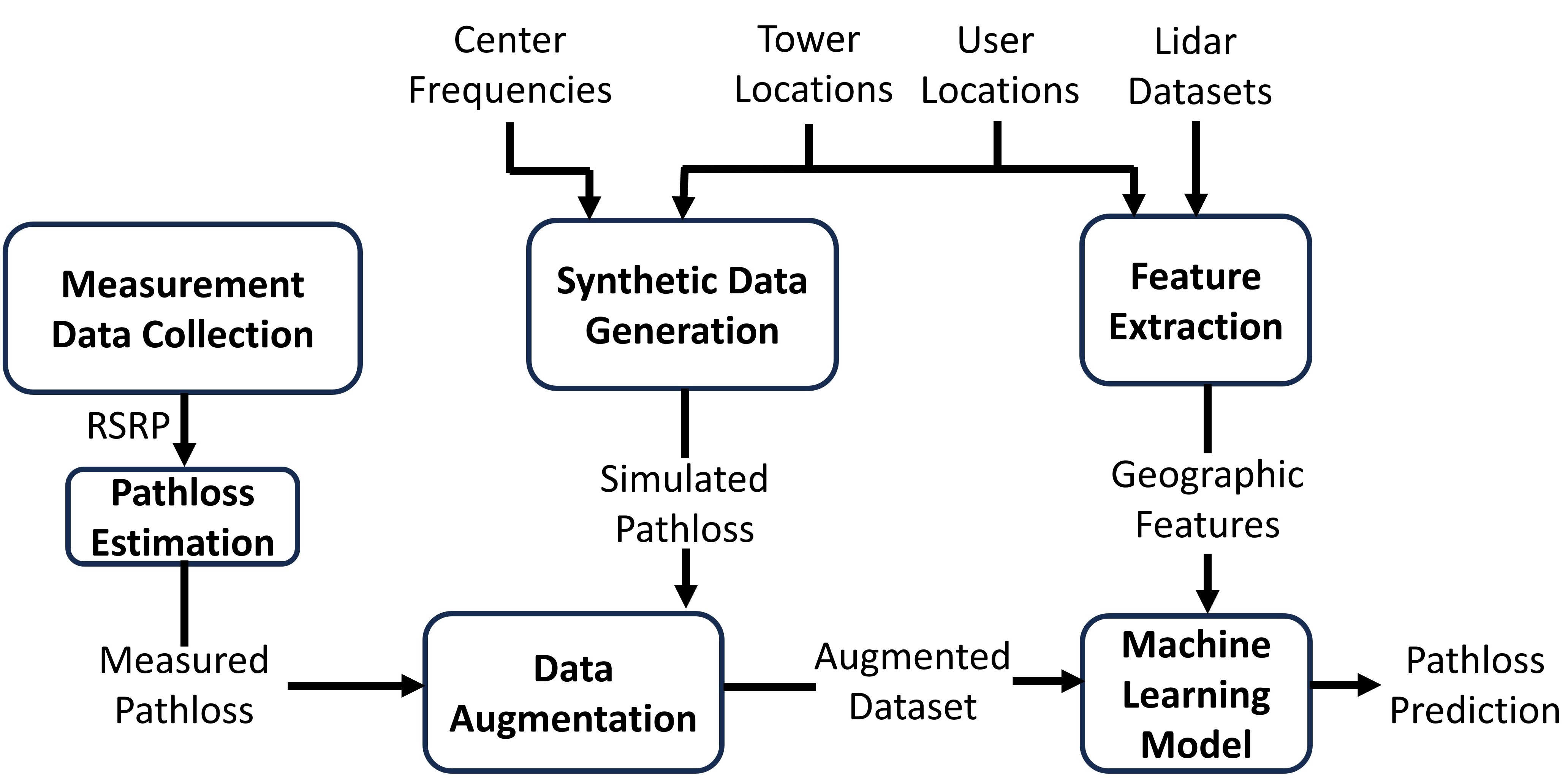}
        \caption{Flowchart of the proposed simulation-enhanced data augmentation process.}
        \label{fig:flowchart}
        \vspace{-3mm}
\end{figure}

In this paper, we introduce a novel data augmentation method for \gls{ml} pathloss prediction, which incorporates both real and synthetic data.
Our goal is to improve the generalizability and reliability of the pathloss prediction model by enriching the dataset with synthetic data, especially in cases where measurement data are limited. 
The proposed data augmentation, as shown in Fig. \ref{fig:flowchart},  has three main components: (i) measurement data collection, (ii) synthetic data generation, and (iii) feature extraction.
For the collection of measurement data, we conducted a measurement campaign in three different environments (rural, residential, and hilly).
We then produce synthetic data using the state-of-the-art large-scale pathloss simulator \cite{zhangLargeScaleCellularCoverage2023}. This simulator utilizes high-resolution LiDAR data to extract features along \gls{los} paths, enabling the prediction of large-scale fading.
We then develop features based on domain knowledge and extract them from LiDAR-based geographic datasets.
We evaluated the prediction performance of the augmented dataset in various scenarios.
The results show that the combination of synthetic and real data can improve the prediction performance for unseen environments at a slight cost of loss of accuracy for known environments.


\section{Simulation-Enhanced Data Augmentation}
Learning-based models often suffer from inaccurate predictions in unfamiliar or unseen environments. 
To address this problem, this paper aims to build a robust model that can perform well not only for known areas but also for unseen scenarios by incorporating synthetic data.
Specifically, we detail the data processing procedure for merging synthetic data with measurements, as well as the data collection procedure.
In this section, we will outline our proposed data augmentation methodology and then describe the procedures for creating both measurement and synthetic datasets.

\subsection{Methodology}
Fig. \ref{fig:flowchart} illustrates the overall flow of the proposed data augmentation.
The three main components of the proposed simulation-enhanced data augmentation can be described as follows:
\begin{itemize}
    \item \textbf{Measurement Data Collection:} 
    Measurement data are collected from mobile phones in different environments.
    These measurements can be used to obtain pathloss values and associated data, such as location, cell information, and carrier frequency.
    \item \textbf{Synthetic Data Generation:} 
    The simulation parameters such as user location and carrier frequency should be determined based on what the measurement dataset lacks, to enhance the dataset.     
    For example, if the dataset lacks rural environment data, simulations for rural environments can be performed using rural geographic data and macrocell parameters.
    \item \textbf{Feature Extraction:}
    After preparing the measurement and synthetic datasets, we extract the features of the data points from the geographic datasets, which are predominantly site-specific. 
    We carefully design features that reflect the geometry and surroundings of the transmitter and receiver using domain knowledge.
    Then, we generate features for every geographic point in the area of interest and pair them with the corresponding pathloss values collected in the previous step.
\end{itemize}



\subsection{Creating Measurement Datasets}
In this section, we will describe how we collected datasets in different environments and the process of converting the \gls{rsrp} measurements into path loss values.
\label{subsec_celldatacollection}
\subsubsection{Collecting \gls{rsrp} Measurements}
\begin{figure*}
     \centering
     \begin{subfigure}[b]{0.32\textwidth}
            {\includegraphics[width=.980\linewidth]{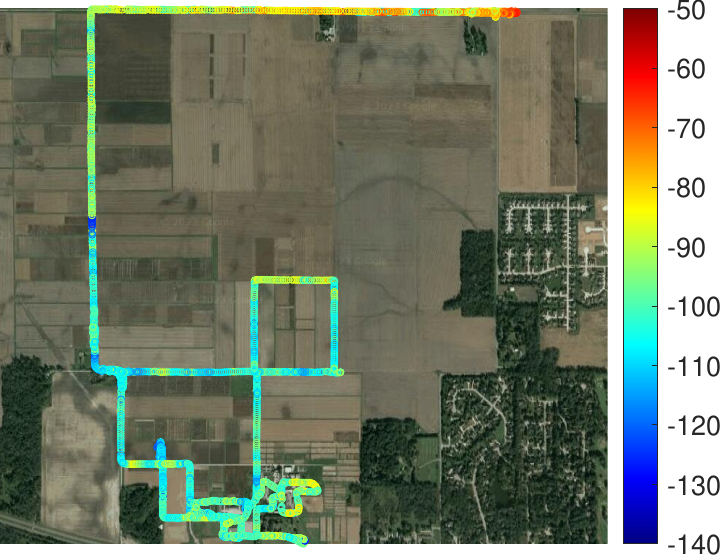}}
            \caption{ Rural (ACRE)}
            \label{fig:ACRE}
     \end{subfigure}
     \begin{subfigure}[b]{0.32\textwidth}
             {\includegraphics[width=.980\linewidth]{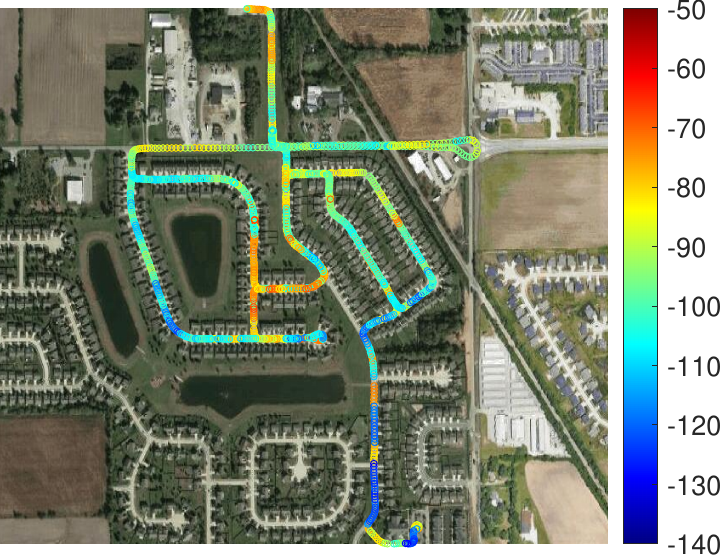}}
             \caption{ Residential (Lindberg)}
             \label{fig:Lindberg}
     \end{subfigure}  
     \begin{subfigure}[b]{0.32\textwidth}
             {\includegraphics[width=.980\linewidth]{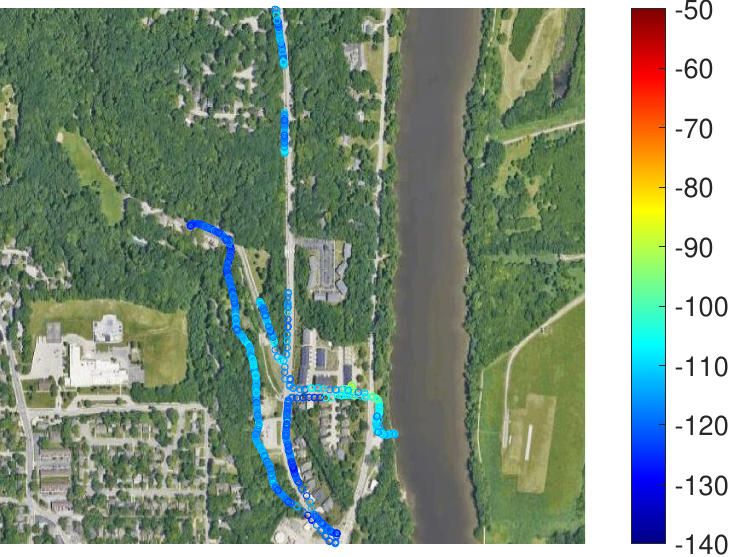}}
             \caption{ Hilly (Happy Hollow)}
             \label{fig:HappyHollows}
     \end{subfigure}   
     \caption{\gls{rsrp} in \si{\dBm} derived from the data collected during the measurement campaign.}
     \label{fig:RSRPmap}
     \vspace{-5pt}
\end{figure*}

We carried out a comprehensive data collection in the city of West Lafayette, located in the state of Indiana in the USA.
We collected data from three different environments: farm, hilly and residential areas, as illustrated in Fig. \ref{fig:RSRPmap}.
We conducted 4G LTE measurements using three commodity Android phones, Samsung Galaxy S8, S20 and S21.
In addition, we used a cellular network monitoring app called G-NetTrack.
Our measurements include \gls{rsrp} measures, cell identifiers (IDs), \gls{earfcn}, and \gls{gps} coordinates.
Note that \gls{earfcn} represents the LTE band and the center frequency of the serving cell.
To acquire cell-site information, we used a mobile application called CellMapper \cite{cellmapperMobilityUnitedStates} and a database known as Antenna Search \cite{AntennaSearchSearchCell}.
CellMapper is a crowd-sourcing application that offers useful cell-specific information, such as cell types, cell IDs, uplink/downlink carrier frequency, and cell addresses.
We determined the addresses of the serving cells using the cell IDs from our measurements. 
However, since CellMapper only provides approximate addresses, we took an additional step using Antenna Search.
Specifically, we searched for the approximate addresses obtained from CellMapper on Antenna Search and extracted detailed tower information such as GPS coordinates and tower height. 

\begin{figure}
    \centering
\includegraphics[width=0.425\textwidth]{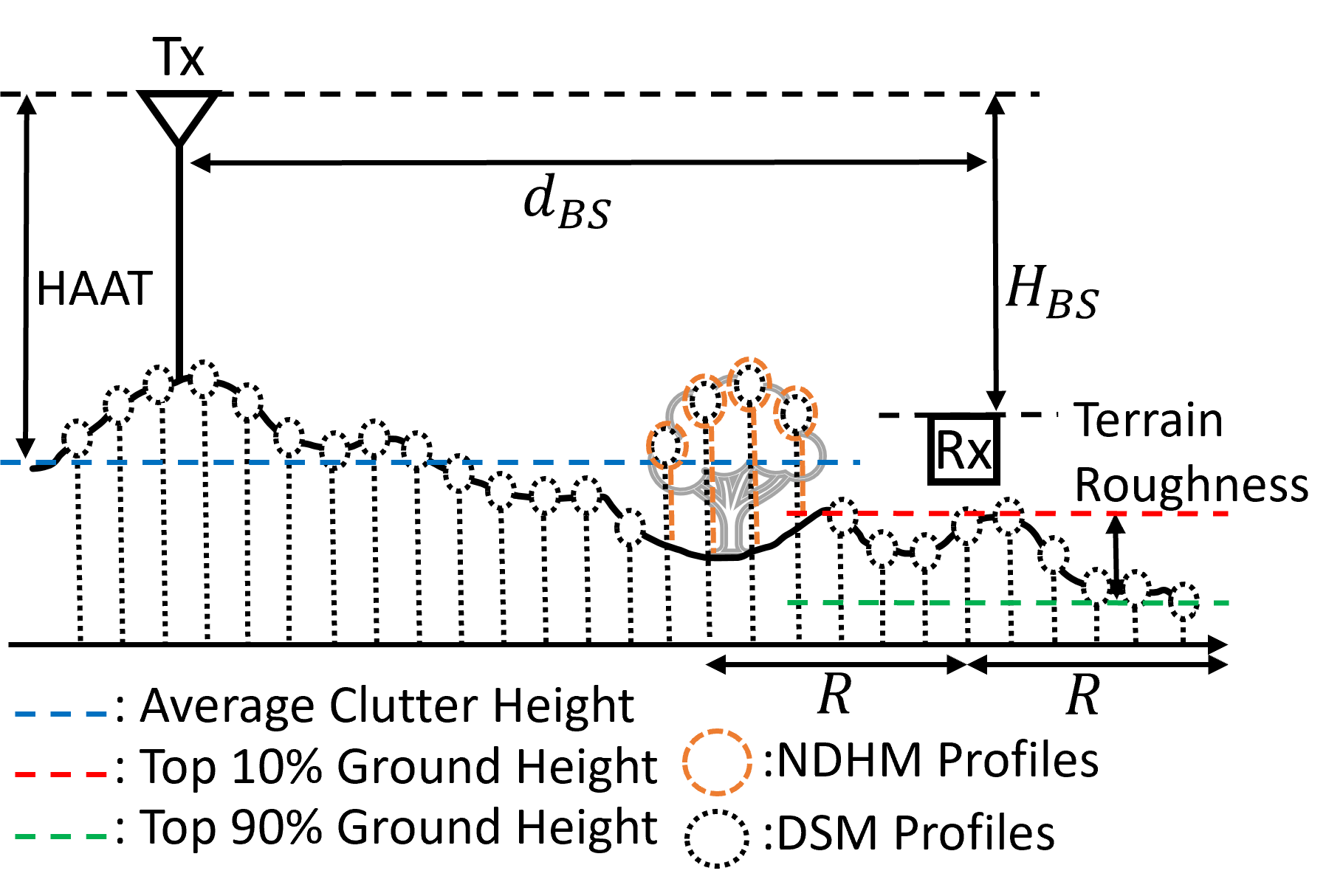}
    \caption{Illustration of the engineered features used as inputs for \gls{ml} algorithm.}
    \label{fig:feature}
    \vspace{-15pt}
\end{figure}

\subsubsection{Converting \gls{rsrp} into Pathloss}
\label{subsec_processingrsrpdata}
In 4G LTE, \gls{rsrp} is the average power received in the resource elements (RE) that carry cell-specific reference signals (CRS).
The \gls{rsrp} in \si{\dBm} can be expressed as \cite{zhangLargeScaleCellularCoverage2023,maengImpact3DAntenna2023}
\begin{equation} 
    \begin{split}
        \gls{rsrp}   
                =-PL+ \Delta \,      
    \end{split}
\end{equation} 
where $PL$ is the pathloss in \SI{}{\deci\bel} and $\Delta$ is an offset.
The offset $\Delta$ represents the cumulative effect of unknown site-specific parameters, including transmit power, antenna gain, and cable loss.
Since these parameters are in general not available, we have to estimate the offset $\Delta$ of each site to compute the path loss.
To this end, we compared \gls{rsrp} in the measurements and pathloss in the synthetic data point-to-point.
The difference between the $\ell$th \gls{rsrp} measure and its corresponding simulated point can be written as $\Delta_{\ell}$.
Then, the offset $\Delta$ can be obtained by taking the average of the differences as
\begin{equation}
    \Delta = \frac{1}{N_s}\displaystyle\sum_{\ell=1}^{N_s}\Delta_\ell,
\end{equation}
where $N_s$ is the number of \gls{rsrp} samples.
In practice, the averaging process can cause additional errors.
Therefore, it would be desirable to derive the exact offsets if site-specific parameter information is available.
Once the offsets are calculated, the pathloss can be derived by subtracting the values of \gls{rsrp} from the offsets, that is, $PL=\Delta-\gls{rsrp}$.

\subsection{Creating Synthetic Datasets}
We utilized a comprehensive cellular coverage simulator~\cite{zhangLargeScaleCellularCoverage2023} to estimate generate pathloss 
for various scenarios, based on 
high-accuracy LiDAR data with meter-level 
resolution. The simulation area for each scenario was designed to cover all relevant measurement points and their surrounding environments. 
Typical LTE network parameters and equipment specifications were applied 
(\Cref{subsec_processingrsrpdata}).

As introduced in \cite{zhangLargeScaleCellularCoverage2023}, two sets of simulation results are available for analysis: pathloss based on the \gls{ehata} model and cumulative blockage distance based on the 60\% clearance test of the first Fresnel zone. The locations of the cell towers were determined by a manual process that involved cross-referencing of data from various publicly accessible tools and sources (\Cref{subsec_celldatacollection}). 
All carrier frequencies for each scenario, as observed in the corresponding measurement data set, were included in the simulation.


\section{Feature Extraction}

In this section, we develop and extract the features of the pathloss prediction model.
Given the significant impact of feature engineering on \gls{ml} prediction performance, we carefully define and extract geographical features using geographical datasets.

\subsection{Geographic Dataset}
To extract geographic features from the LiDAR datasets, we used Indiana's statewide LiDAR datasets collected in 2018 and \gls{dsm} and \gls{dhm} for geographic profiles \cite{PURR3707}.
The \gls{dsm} provides 
the elevation above sea level for each point, representing the vertical height of the tallest objects, including trees and buildings at those points.
In contrast, \gls{dhm} provides information about the elevation above ground level, which indicates the height of the clutter (e.g., buildings and trees) at each point.
The \gls{dsm} and \gls{dhm} have a spatial resolution of 5 feet.
With these two datasets, we can determine the elevation of the ground and the height of the surface of each point.

\subsection{Feature Engineering}
For radio attributes, we focus on the center frequency, given its significant influence on pathloss \cite{phillipsSurveyWirelessPath2013}. 
In terms of geographic attributes, we have incorporated both endpoint- and path-based characteristics, including the relative height of the serving cell to each reception point and the elevation angle. 
Fig. \ref{fig:feature} illustrates the radio and geographical features.
The description of the features is given below.
\begin{itemize}
    \item \textbf{Carrier Frequency:}   Represents the center frequency at which the communication signal is transmitted. 
    This parameter can be extracted directly from the attribute \gls{earfcn} in the measurement dataset.

    \item \textbf{Serving \gls{bs} Distance ($d_{BS}$):} 
    Defined as the distance between a receiving point and its service \gls{bs}.

    \item \textbf{Relative \gls{bs} Height ($H_{BS}$):} 
    The height of the serving \gls{bs} relative to a receive point.
    
    \item \textbf{Average Clutter Height ($H_C$):} Average relative height of clutter of neighboring points.
    Neighboring points are defined as points within a radius circle $R$ centered on a received point.
    In this study, we choose to use $R=\SI{50}{m}$.
    
    \item \textbf{Terrain Roughness:} 
    This parameter characterizes the roughness of the terrain, which is defined as the difference between the top $10\%$ and $90\%$ ground elevation of the neighboring points. 
    
    \item \textbf{Transmitter Height Above Average Terrain (TxHAAT):} 
    The difference between the \gls{bs} height and the average clutter height.
    This parameter provides information about the height of \gls{bs} above the surrounding terrain.
    
    \item \textbf{Ratio $\bm{\alpha}$} \cite{reus-munsMachineLearningbasedMmWave2022}: 
    The ratio $\alpha$ is defined as $\frac{H_{BS}-H_{C}}{d_{BS}}$. This ratio is used to characterize the elevation angle between the serving \gls{bs} and a receiving point.

\end{itemize}


\section{Performance Evaluation}
For the performance evaluation, collected measurements were used from three different areas. The collected datasets are composed of about 133,800 \gls{rsrp} measurements, spanning three different scenarios:
\begin{itemize}
\item \textbf{ACRE} (71,068 samples): Rural farm area
\item \textbf{Lindberg} (16,107 samples): Residential area.
\item \textbf{Happy Hollow} (46,641 samples): Hilly suburban area 
\end{itemize}


We created simulations for ACRE, Lindberg, and Happy Hollow based on the cell information acquired from the measurement campaign.
We configured user location grids with 6300 points for ACRE, 9200 points for Lindberg, and 5900 points for Happy Hollow to fully cover the area of interest.
We simulated each site with varying center frequencies from \SI{731.5}{\mega\hertz}  to \SI{2538.2}{\mega\hertz} to account for different propagation characteristics.
In the end, we obtained a synthetic dataset of approximately 651,000 data points, which improved our training datasets with different environments.


To evaluate the prediction performance of the \gls{ml} models, we use \gls{mae}, which is expressed as
\begin{equation}
MAE = \frac{1}{n} \sum_{i=1}^{n} |L_i - \hat{L}_i|
\end{equation}
where $n$ is the number of pathloss samples, $L_i$ is the true pathloss, and $\hat{L}_i$ is the predicted pathloss.

\begin{figure}
    \centering
\includegraphics[scale=0.1]{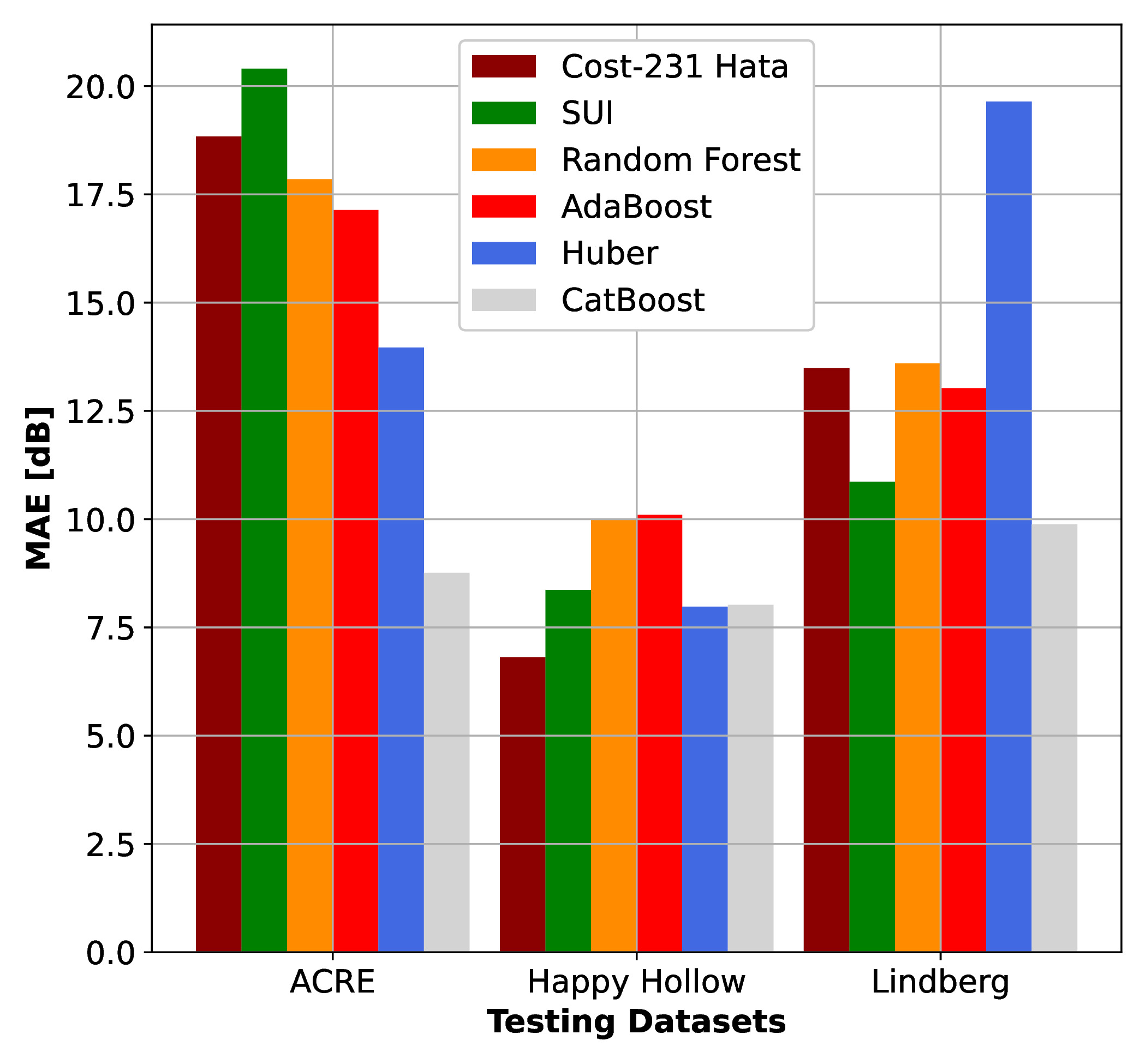}
 \caption{Prediction performance in the same environment, where training data consists of synthetic data structured to mimic the characteristics of the real test data environment.}

    \label{fig:modelComp}
    \vspace{-6mm}
\end{figure}


\begin{table*}
    \centering
    \begin{tabular}{@{}lcc@{}}
        \toprule
        Training Data & \multicolumn{2}{c}{MAE [dB] for Different Testing Datasets} \\
       \midrule
        Scenario 1 & ACRE& Happy Hollow\\
        \midrule
        \rowcolor{lightergray} ACRE (R) + Happy Hollow (S) & $\textbf{4.97}$ & $\textbf{7.7}$ \\
        ACRE (R) & $4.33$ & $8.7$ \\
        Happy Hollow (S) & $8.6$ & ${7.98}$ \\
        \midrule
        \rowcolor{lightergray} Happy Hollow (R) + ACRE (S) & $\textbf{8.74}$ & $\textbf{5.55}$ \\
        Happy Hollow (R) & $9.65$ & $5.22$\\
        ACRE (S) & $8.74$ & $6.59$ \\
        \midrule
        Scenario 2 & Lindberg& ACRE\\
        \midrule
        \rowcolor{lightergray} Lindberg (R) + ACRE (S) & $\textbf{6.32}$ & $\textbf{8.75}$ \\
        Lindberg (R) & $5.39$ & $10.49$ \\
        ACRE (S) & $10.14$ & $8.74$ \\
        \midrule
        Scenario 3 & Lindberg  & Happy Hollow\\
        \midrule
        \rowcolor{lightergray} Lindberg (R) + Happy Hollow (S) & $\textbf{9.5}$ & $\textbf{7.79}$ \\
        Lindberg (R) & $5.39$ & $19.99$ \\
        Happy Hollow (S) & $10.23$ & ${7.98}$ \\
        \bottomrule
    \end{tabular}
    \caption{    
    Pathloss prediction accuracy (in MAE) comparing the proposed data augmentation (highlighted) to baselines. "R" denotes real data and "S" denotes simulated data. For each dataset, training includes all synthetic data and 50\% of the real data, the remaining real data being used for evaluation.
    }
    \label{tab:general}
    \vspace{-15pt}

\end{table*}

\subsection{Comparison with Empirical Radio Propagation Models}
We compared our ML-based pathloss prediction with traditional empirical models\footnote{The source code for this work is available publicly at https://github.com/aprincemohamed/DeepLearningBasedCellularCoverageMap
.}
 that use an equation to calculate pathloss at any given location from the base station \cite{masood2019machine}.
Specifically, we used the COST-231 Hata model and the Stanford University Interim (SUI) model as benchmarks \cite{abhayawardhana2005comparison}.
The COST-231 Hata model includes adjustments for different environments, such as urban, suburban, and rural areas.
The SUI model refines the predictions by offering correction factors for antenna heights of the base station and user equipment, along with constants that vary depending on the type of terrain.
We applied terrain A corrections for the Lindberg and Happy Hollow datasets and terrain C corrections for the ACRE dataset, as specified for the SUI model \cite{abhayawardhana2005comparison}.

 Fig. \ref{fig:modelComp} compares the prediction performance of the deterministic and \gls{ml} models at different sites. 
We evaluated several \gls{ml} algorithms, including Random Forest, AdaBoost, HuberRegressor, and CatBoost, to identify the one that offers the highest accuracy. 
The models were trained on a synthetic dataset and then validated against a real-world dataset.
Specifically, for the Happy Hollow datasets, excluding the center frequency feature from training and testing improved the model's performance over using all features.

In general, it can be verified that CatBoost outperforms the other \gls{ml} schemes and empirical models. 
The MAE achieved for the ACRE, Happy Hollow, and Lindberg data sets was approximately \SI{8.76}{\deci\bel}, \SI{8.02}{\deci\bel}, and \SI{9.88 }{\deci\bel}, respectively. 
It should be noted that synthetic data, coupled with the developed features, contributed significantly to favorable results. 
This is evidenced by consistently lower MAE values, which significantly undercut those obtained with empirical models. 
For the Happy Hollow dataset, the COST-231 Hata model achieved an MAE of \SI{6.8}{\deci\bel}, which is marginally better than the \SI{8.02}{\deci\bel} of CatBoost when using synthetic data. 
However, when CatBoost is trained on real data, specifically with a 50\% split, it exhibits superior performance with an MAE reduced to approximately \SI{5.22}{\deci\bel}, outperforming the COST-231 Hata model.
We infer that CatBoost's superior performance can be attributed to its sophisticated ordered-boosting technique, which is especially effective in avoiding overfitting.
Hence, we chose CatBoost as our main ML scheme for the following experiments.

These comparisons clearly highlight the advantage of using machine learning over empirical models. 
This approach circumvents the need for costly data collection processes by showing that our models, when trained with cost-effective synthetic data, still maintain high performance in real-world scenarios. 
This is particularly advantageous because it is achieved without reliance on complex or expensive data types, such as images, highlighting the practicality and efficiency of our methodology.

\subsection{Generalization Performance Evaluation}
In Table \ref{tab:general}, we assess how effectively our proposed training method generalizes across different scenarios, specifically comparing its performance when using only measurement data versus scenarios that use only synthetic datasets. For the measurement data at each location, we partitioned the dataset into two equal halves, allocating 50\% for training purposes and the remaining 50\% for testing. Generally, cases that use only measurement data show good prediction accuracy in known areas; however, their performance decreases in unseen areas. In all scenarios, the cases that used only measurement data showed lower performance compared to those that only used simulation data to predict unseen areas.
It can be seen that the proposed training method improved the accuracy of the prediction of unseen areas by incorporating synthetic data, although at the cost of some accuracy in known areas. 
Specifically, the ACRE (R) + Happy Hollow (S) case improved the prediction accuracy of Happy Hollow by \SI{1.0}{\deci\bel} compared to the ACRE (R) case. On the other hand, the accuracy of the ACRE prediction decreased slightly by \SI{0.64}{\deci\bel}.
Similarly, the Happy Hollow (R) + ACRE (S) case improved the prediction accuracy of ACRE by \SI{1.09}{\deci\bel}, with a minor loss of the prediction accuracy of \SI{0.33}{\deci\bel} for Happy Hollow.
A similar trend can be observed in Scenario 2.
The proposed data augmentation improved the prediction accuracy for ACRE by \SI{2.26}{\deci\bel} at the cost of a \SI{1.07}{\deci\bel} accuracy loss for Lindberg.

In Scenario 3, the only real data-based scheme achieves the prediction accuracies of \SI{19.99}{\deci\bel} and \SI{5.38}{\deci\bel} for Happy Hollow and Lindberg, respectively.
The large difference in the prediction results for Happy Hollow (hilly) and Lindberg (residential) may suggest that they have significantly different propagation environments. 
However, when the synthetic data for Happy Hollow was added to the dataset, the prediction accuracy for Happy Hollow was enhanced by approximately \SI{12.2}{\deci\bel}.
This improvement came with a trade-off, leading to a loss \SI{4.11}{\deci\bel} in the accuracy of the Lindberg prediction, which is consistent with previous findings.

These results suggest that the incorporation of synthetic data into training improves prediction accuracy in almost all test scenarios, as seen in the lower dB values for the combinations of real and synthetic data compared to real data alone or synthetic data alone. 
This aligns with our earlier discussion on the potential advantages of leveraging synthetic data to address the generalization challenge posed by data scarcity.



\begin{figure}
    \centering    
    \includegraphics[scale=0.08]{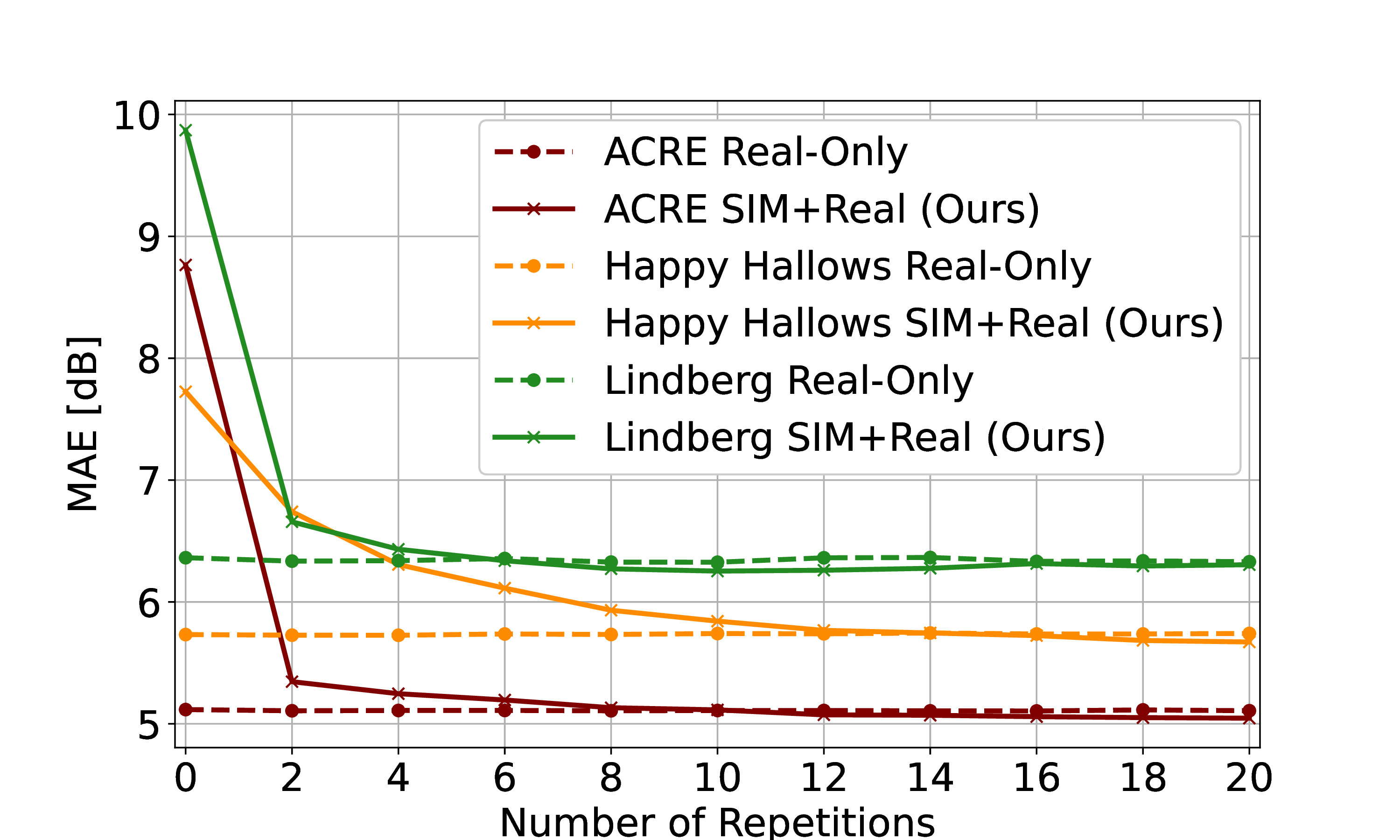}
    \caption{MAE vs 5\% Real Data Repetitions in Training Set.}
    \label{fig:sim&real}
    \vspace{-18pt}
\end{figure}
\subsection{Training Optimization with Limited Measurement Data}
Next, we evaluate the prediction performance when available measurement data is limited.
In real-world applications, the challenge of data scarcity is common. 
To navigate this, we use a nominal 5\% of available measurement data in the training process.
To evaluate the impact of simulation-aided data augmentation, we compared our proposed method with a baseline that uses only measurement data. 
In our approach, for each site, we randomly sampled only 5\% of the entire measurement dataset for training and used the remaining $95\%$ for testing.
To counterbalance the limited measurement data, we augment it with synthetic data, exposing the model to a broader range of scenarios. 
The synthetic dataset, which is significantly larger than the real data, risks overfitting the model to its characteristics. To address this issue, we implemented a strategy of repeating the measurement dataset multiple times in each epoch, ensuring a balanced distribution of real and synthetic data during the model training phase.

Fig. \ref{fig:sim&real} shows the prediction accuracy for different scenarios with varying numbers of repetitions.
Clearly, our approach, which involves repeated exposure to a small subset of measurement data, results in significant improvements in predictive accuracy.
Specifically, for ACRE datasets, employing 20 repetitions of the 5\% real data subset, the model achieved an MAE of \SI{5.05}{\deci\bel}, a significant improvement over the MAE of \SI{8.77}{\deci\bel} observed when the model was trained only using synthetic data.
Interestingly, the figure also highlights the benefits of our repetitive method compared to using only real data without repetition, which yielded a slightly higher MAE of \SI{5.11}{\deci\bel} using 16 repetitions. 
Similarly, for the Happy Hollow site, the MAE was reduced to \SI{5.67}{\deci\bel} with 20 repetitions, compared to \SI{7.72}{\deci\bel} using only synthetic data and slightly better than \SI{5.74}{\deci\bel} observed with 18 repetitions of only real data.
Furthermore, for the Lindberg dataset, where the model reported an MAE of \SI{6.25}{\deci\bel} with 10 repetitions, significantly outperforming the MAE of the synthetic data of \SI{9.87}{\deci\bel}, and also showing a slight advantage over the MAE of \SI{6.32}{\deci\bel} achieved with 10 repetitions of only real data.

By addressing data imbalance, we improved the prediction accuracy of our model, even with limited 
measurements in training sets. This underscores the efficacy of our simulation-aided data augmentation, which proves its potential as a solution to 
measurement scarcity in real-world applications. Our approach enhances performance beyond what is achieved with either synthetic or real data alone. It showcases the advantages of a balanced mix of both, thus optimizing predictive accuracy across various datasets.

\section{Conclusion}

This paper addressed a simulation-enhanced data augmentation method to overcome the data shortage problem in the prediction of pathloss via \gls{ml}.
We collected data for three different environments using commodity mobile phones and augmented the dataset using synthetic data to enhance diversity.
For the \gls{ml} prediction model, we engineered and extracted geographic features from LiDAR datasets.
We conducted a rigorous performance evaluation to show the effectiveness of the proposed method.
The experimental results demonstrated that the proposed data augmentation method effectively enhances the generalization performance of the prediction model.
In summary, our research offers a practical and effective strategy for the prediction of pathloss, highlighting the potential of simulation-aided \gls{ml} to mitigate data scarcity challenges and improve the efficiency of network deployment.


\bibliographystyle{IEEEtran}
\bibliography{IEEEabrv,references}

\end{document}